\documentclass[journal]{IEEEtran}

\ifCLASSINFOpdf
\else
   \usepackage[dvips]{graphicx}
\fi
\usepackage{url}

\usepackage{comment}
\usepackage[dvipsnames]{xcolor}
\usepackage[normalem]{ulem} 
\usepackage{graphicx}
\usepackage{caption}
\usepackage{subcaption}

\begin{document}

\title{Does Optimal Source Task Performance  Imply Optimal Pre-training for a Target Task?}

\author{Steven Gutstein, Brent Lance \IEEEmembership{Senior Member, IEEE}, and Sanjay Shakkottai \IEEEmembership{Fellow, IEEE}
\thanks{This work was supported in part by the U.S. DevCom Army Research Laboratory under ARL-AY6-04 and Army Research Office Grant W911NF-17-1-0359.}
\thanks{Steven Gutstein and Brent Lance are with the Army Research Laboratory, 
Aberdeen Proving Ground, MD 21005 (e-mail: steven.m.gutstein.civ@mail.mil; 
brent.j.lance.civ@mail.mil)}
\thanks{Sanjay Shakkottai is with the Dept. of Electrical and Computer 
Engineering at the University of Texas at Austin, Austin, TX 78712 USA (e-mail: 
sanjay.shakkottai@utexas.edu).}}

%\markboth{IEEE Transactions on Neural Networks and Learning Systems, Vol. ??, No. ?, April 2021}

\maketitle

\begin{abstract}
Fine-tuning of pre-trained deep nets is commonly used to improve accuracies and training 
times for neural nets. It is generally assumed that pre-training a net for optimal source task performance best prepares it for fine-tuning to learn an arbitrary  target task. This is generally not true.  Stopping source task training, prior to optimal performance, can create a pre-trained net better suited for fine-tuning to learn a new task. We perform several experiments demonstrating this effect, as well as the influence of the amount of training and of learning rate. Additionally, our results indicate that this reflects a general loss of learning ability that even extends to relearning the source task.
\end{abstract}


\IEEEpeerreviewmaketitle

\section{Introduction}

Fine-tuning of pre-trained nets is a standard transfer learning approach for deep nets. It entails 
training a net on a source task, randomly reinitializing one or more of the higher layers, and then retraining the resultant net on a desired target task. This reasoning is based on the structure of the network, namely, layers closer to the input extract general features whereas layers closer to the output extract more task/dataset specific features \cite{YosinskiCBL14}. 

The advantages of reusing general features are 
three-fold: 1) Better training times, 2) Better accuracy, and 3) Reduced data requirements.

The logic of pre-training deep neural networks and its success in a wide variety of domains have led to its routine use. State-of-the-art (SOTA) results have been achieved for object detection, image segmentation and other recognition tasks \cite{GirshickDDM13, Girshick15, LongSD14, SimonyanZ14,CarreiraZ17, Sermanet2013overfeat} by using nets that have been pre-trained on the ImageNet dataset \cite{Deng09} 
as the backbone of the net to be trained on the target task.

Investigations of the methodology for pre-training followed by fine-tuning have generally focused 
either upon which layers to transfer \cite{YosinskiCBL14}, or on using source task training sets significantly larger than ImageNet, such as ImageNet-5K \cite{HeGDG17}, JFT-300M dataset \cite{SunSSG17} and Instagram \cite{Mahajan18}. They do not focus on how, for a given source task, to best pre-train a net for an arbitrary transfer task.

In this paper we  illustrate a new consideration for pre-training - the amount of training performed on the source task. It has recently been demonstrated that, in many cases, target task datasets must be far smaller than  previously believed before any accuracy benefits accrue from pre-training \cite{He18}. Therefore, our primary focus is on small target task training sets with 10 Samples per Class(SPC). However, we also examine larger target task training sets of 50 SPC and 250 SPC.

A standard assumption of pre-training with fine tuning is that training a net to peak performance on a source task best prepares it to learn a target task. This overlooks the possibility of overtraining (i.e. learning source task specific features that do not generalize to the target task) in a manner analogous to the way training too much on a training set will decrease a net's generalizability to an arbitrary test set. It also overlooks the possibility that a net's ability to respond to new training data may decrease. 

Recent work on incremental learning \cite{achille2017critical, shwartz2017opening, chaudhry2018riemannian,  saxe2019information, ash2020warm, berariu2021study} indicates that learning in deep nets is a 2 phase process. The first phase involves rapidly acquiring information in a ``memorization" stage; the second involves pruning unnecessary connections in a ``reorganization" phase. Although classification performance continues to improve during the ``reorganization" phase, the net's ability to respond to training data decreases. Achille et al.\cite{achille2017critical} refer to this as a loss of ``Information Plasticity".

The implications of this for pre-training with fine tuning are clear. The benefit of information gained from the source task may be paid for with a loss of learning ability. Reinitialization of the upper layer(s) will restore some Information Plasticity, but it is not clear how much. This raises the question of how to best balance these two factors.

%Although reinitialization of the upper layer(s) should restore some of a net's ability to respond to training data, will this compensate for the amount of training required to achieve peak source task performance, or would better target task performance be achieved with sub-optimal source task performance?

The contribution of this paper is to use the CIFAR100 \cite{Krizhevsky09}, 
Tiny ImageNet200 \cite{Li17} and CalTech101 \cite{Fei04} datasets with both a Wide ResNet-28-10 (WRN-28-10)\cite{Zagoruyko16} and a VGG16 architecture \cite{simonyan2014very} to demonstrate that:

\begin{itemize}
    \item Optimal source task performance does not imply optimal transfer to a target task
    \item This reflects a general loss of learning ability, rather than overtraining on the source task.
\end{itemize}

%The contribution of this paper is to use the CIFAR100 \cite{Krizhevsky09}, 
%Tiny ImageNet200 \cite{Li17} and CalTech101 \cite{Fei04} datasets with both a Wide ResNet-28-10 (WRN-28-10)\cite{Zagoruyko16} and a VGG16 architecture \cite{simonyan2014very} to demonstrate that:
%\begin{itemize}
%	\item Too much source task training (i.e. pre-training) can significantly degrade a pre-trained net's ability to learn a target task with fine-tuning.
%	\item This loss reflects a general loss of learning ability. It has the characteristics of `fragile co-adaptation', which Yosinski et al. \cite{YosinskiCBL14} argue can leave a pre-trained net less able to relearn a source task or learn a related task. 
%	\begin{itemize}
%     	\item   We are the first to connect this effect with the amount of source task training, rather the layers selected for reinitiaization.
%	    \item Unlike Yosinski et al. \cite{YosinskiCBL14}, we observed this when reinitializing only the uppermost layer of the net, rather than  several of the top layers. This observation is important
%	\end{itemize}
	
%\end{itemize}

\section{Related Work}
Two of the main hurdles in training neural nets are the large quantities of time and data required. These can be mitigated with transfer learning, which allows a deep net to reuse weights learned from a source task to more rapidly learn a target task. In spite of the lack of explainability of deep nets, their structured nature lends itself to transfer learning. It has long been known that deep neural nets trained on visual data learn very similar first layer features - namely Gabor features and color patches, regardless of the specific task they have been trained upon \cite{Le11, Krizhevsky12, Lee09}. Yet, the output layers of nets are, by necessity, extremely task specific.  It seems intuitively obvious that initializing a net with features that are more similar to the desired end state should save training time and allow for improved accuracy with smaller training sets. 

The technique of copying lower layers of a net and training higher layers from random initialization is known as `pre-training and fine-tuning'. Indeed, as noted in \cite{He18}, the first significant results using deep learning approaches for detecting objects involved pre-training with the ImageNet dataset \cite{Deng09}  before learning the target (i.e. desired) task \cite{Girshick15, Sermanet2013overfeat, He18, Oquab14}. This is now the dominant approach for state of the art object detectors \cite{Girshick15,HeGDG17, RedmonDGF15, Wu19,LiuAESR15,LinDGHHB16}

It is possible to view pre-training as a form of initialization. Determining the best way to initialize a network to learn was a subject of early research. For instance, initializing all the weights of a net to a constant is poor practice, since the net must somehow break symmetry if it is to learn. This led to the development of several initial randomization techniques \cite{lecun2012efficient, Glorot10a, saxe2013exact, He_2015_ICCV}, each of which was intended to make the net as receptive as possible to the desired training. 

Recent research has found that as a net trains, it eventually becomes less sensitive to training data. For example, the common technique of using a ``warm-start" for training (i.e. updating a model with new data, rather than retraining it from scratch with the combined new \& old data), results in a model with generalization inferior to training with the complete data set \textit{ab initio} \cite{ash2020warm, berariu2021study}. These results are consistent with a \textit{two phase} conception of learning introduced by Achille et al.\cite{achille2017critical} and Shwartz and Tishby \cite{shwartz2017opening}. Using the Fisher Information of the network weights to measure the effective connectivity between layers of a network during training, Achille et al.\cite{achille2017critical} observed that initially, information rises rapidly before decreasing, preventing redistribution of information resources. They defined this effect as a loss of “Information Plasticity”.

Pre-training with fine tuning entails a reinitialization of a net's upper layers, which could conceivably restore some Information Plasticity. However, earlier work on pre-training and fine tuning by Yosinski et al. \cite{YosinskiCBL14}, which was aimed at localizing the layers of a net responsible for transfer learning, found that keeping the lower and middle layers led to worse performance than keeping just the lower layers, or keeping all layers save the output layers. 

This was attributed to a `fragile co-adaption' of the nodes in the middle layers, which they explained by stating that the net's layers learned ``features that interact with each other in a complex or fragile way such that this co-adaptation could not be relearned" \cite{YosinskiCBL14}. In other words, as a net trains, each layer learns its weights in conjunction with the changing outputs of the layer beneath it. It can be very difficult for Stochastic Gradient Descent (SGD) to rediscover those same final weights, or functionally equivalent ones, from a random initialization, even when all the layers beneath it are in the state for which those weights were learned.

This leads to our question, whether there is some degree of pre-training epochs that gains enough information about a source task to be useful, while still retaining enough Information Plasticity to allow learning a target task.
%This leads us to ask the question: how can one best 
%pre-train? Previous studies have centered on which layers to transfer and how 
%large the transfer training set should be \cite{YosinskiCBL14,GutsteinFF07}. It 
%has also been observed that the accuracy benefits of transfer learning decline 
%rapidly with increasing size of the target task training 
%set \cite{He18, GutsteinFF11}. 

%In this paper we ask: how much pre-training should be performed? 
%Yosinski et al. \cite{YosinskiCBL14} had observed that keeping the lower and middle layers led to worse performance than keeping just the lower layers, or keeping all layers save the output layers. They attributed this to a `fragile 
%co-adaption' of the nodes in the middle layers. They explain this term by stating that the 
%net's layers learned ``features that interact with each other in a complex or 
%fragile way such that this co-adaptation could not be 
%relearned" \cite{YosinskiCBL14}. In other words, as a net trains, each layer learns its weights in conjunction with the changing outputs of the layer beneath it. It can be very difficult for Stochastic Gradient Descent (SGD) to rediscover those same final weights from a random initialization, even when all the layers beneath it are in the state for which those weights were learned. We observed a similar effect at the top layers of the net, which manifested  after too much source task training. Because this effect manifested at top network layers, it may impact practitioners who rely upon using fine-tuning with pre-trained nets.

\section{Experiments}
\subsection{Datasets}
We performed experiments using three standard datasets:
%\begin{enumerate}
1) CIFAR100 \cite{Krizhevsky09}, 2) Tiny ImageNet200 \cite{Li17} and 3)  Caltech101 \cite{Fei04}. CIFAR100 \cite{Krizhevsky09} contains 100 classes divided into a training set with 50,000 images (500 Samples per Class (SPC)) and a test set with 10,000 images (100 SPC). All images in CIFAR100 were 32x32x3.
 
 Tiny ImageNet200 \cite{Li17} consists of 200 classes drawn from the ImageNet dataset. It also had 500 SPC in the training set and 100 SPC in its test set. However, the images in this set were 64x64x3. Hence, we downsampled to 32x32x3 both for consistency and to decrease training time. 

 Finally, the Caltech101 dataset \cite{Fei04} is comprised of 101 samples with SPC's ranging from a few dozen to a few hundred. On average each class had about 50 SPC. We similarly resized the Caltech101 images to 32x32x3 to decrease training time and for consistency, since they were not all of uniform size. 
 
 %\subsubsection{Division into Source and Target Classes}
 Each dataset was divided into a sets of living and not living classes. This was done to ensure that there would not be overly similar classes in both the source and target sets and to ensure  source and target tasks with an appreciable number of classes.

The CIFAR100 dataset was divided into a source task of 65 living classes, and a target task of 35 not living classes. The Tiny ImageNet200 dataset was divided into a source task of 65 not living classes and  a target task of 35 living classes. Finally, the Caltech101 dataset was divided into a source task of 51 living and a target class of 49 not living classes. These splits were determined by class availability and the desire to have more classes in the source tasks than in the target tasks.

\subsection{Neural Net Architecture}
We used two different deep net architectures. The first architecture we used was a WRN-28-10 \cite{Zagoruyko16} architecture. We chose it because it achieved high performance on the CIFAR100 dataset without requiring novel augmentation or regularization techniques, balancing good performance with ease of implementation. Because we are using Keras (v. 2.3.0), but Zagoruyko et al.'s \cite{Zagoruyko16} original implementation used  PyTorch, we based our instantiation on the Keras based code at: https://github.com/titu1994/Wide-Residual-Networks.

The second architecture we used was VGG-16 \cite{simonyan2014very}. This was chosen because it's a standard deep architecture, which also achieved reasonable performance on CIFAR100, and trained significantly more quickly than the WRN-28-10. We based our instantiation on code at https://github.com/geifmany/cifar-vgg.

\subsection{Training Procedure}
All experiments were run using Keras (v. 2.3.1) with a Tensorflow backend (v. 2.1.0) on machines with Quadro M6000 GPU's and either CUDA 10.2 or 11.0. Minibatches of 128 images were preprocessed with featurewise centering and normalization. Also, data augmentation,  using both random horizontal image flips and uniformly random vertical \& horizontal shifts of up to 4 pixels, was employed.  Both of these techniques were implemented with the Keras image preprocessing module. After each training epoch, the image order was randomly reshuffled. 

For the WRN-28-10 architecture, we followed Zagoruyko et al. \cite{Zagoruyko16} and used SGD with Nesterov momentum to optimize, with a Categorical Cross-Entropy loss function. We  also used the hyper-parameters they employed for CIFAR100. Most notably, we used their learning rate schedule, which is reflected in the abrupt 
changes the accuracies achieved on the source task validation set at the 60$^{th}$ and 120$^{th}$ epoch, though not by the 160$^{th}$ epoch. For the VGG16 architecture, we maintained the hyper-parameters and training schedule from https://github.com/geifmany/cifar-vgg.

\subsubsection{Source Task Training}
Source task training consisted of training  a net from scratch, using the full source task training set. It lasted for 200 epochs, by which time the net displayed very little change in its validation set accuracy. The state of the net was recorded every 10 epochs, resulting in an initial set of 21 nets (i.e. Epoch 0 - Epoch 200, inclusive) for transfer. Furthermore, we saved the net with the highest validation set accuracy, since it is the one normally selected for transfer. 

This was done 5 times, providing 5 different source nets at each of 21 different recorded source task training epochs and at the optimal source task epoch.

\subsubsection{Target Task Training}
The initial state of the target task net was a trained source task net with the final fully connected layer and softmax layer reinitialized and changed to reflect the new number of output classes.

Most experiments used target task training sets with 10 samples per class (SPC). This is because the accuracy benefits of transfer learning are most visible with small target task training sets, so transfer learning is more important for small training sets. This also enabled us to choose 5 different target task training sets using random sampling without replacement for each target task training set. By combining them with the 5 different source nets, we performed 25 quasi-independent transfer experiments both for each source task training epoch and for each optimal source task epoch. 

In addition to the 10 SPC experiments, we also performed experiments with 50 SPC and 250 SPC using the CIFAR100 dataset, so that we could observe how sensitive larger target training sets were to the amount of source task training. However, as we increased the SPC in the target task training sets, we increased the degree of overlap between different target task training sets, since the complete target task training set contained only 500 SPC.

Our graphs reflect the minimum, median and maximum accuracy achieved on the target class validation set. This set was the same for all transfer experiments within each dataset. 

\subsection{Specific Experimental Questions}
\subsubsection{ Does optimal source task performance imply optimal pre-training for a target task?}
Our initial experiments were designed to illustrate both that training for optimal source task performance might not be optimal for learning a target task, and the degree to which it might be suboptimal.

\subsubsection{Does too much source task training place the pre-trained net in a state that is overly specific to relearning the source task?}
The premise of transfer learning via pre-training is that the weights of the retained layers create an initial state that is conducive for learning tasks similar to the source task. Presumably, the better the original net is at its source task, the better  the resulting pre-trained net will be for learning related tasks. More assuredly, the better the pre-trained net would be at relearning the original source task, since that is the exact task the retained layers had been trained to do. A drop in accuracy when learning a different target task, but not when relearning the original source task, would indicate that the pre-trained net's state had become overly specific for relearning the source task.

To test this, we performed experiments with identical source and target tasks.  If the remaining layers of the net are overly specific for learning the source task, then relearning it should not display a drop in accuracy associated with too many `pre-transfer' source task training epochs, though other target tasks (i.e. different than the source task) might display such a drop. If, however, this effect is present, then we are observing a loss of ``Information Plasticity" similar to what Yosinski et al. \cite{YosinskiCBL14} saw and ascribed to  ``fragile co-adapted features"; essentially, the net's upper layer(s) are reinitialized, it is able neither to relearn the source task, nor to learn related tasks.  

\subsubsection{Does target task training set size affect loss of transfer accuracy associated with too much source task training?}

We performed experiments using target task training set sizes of 10, 50 and 250 SPC with the CIFAR100 dataset, using both the WRN-28-10 and VGG16 architectures. Because the benefits of transfer learning decrease with increasing target class size \cite{He18, GutsteinFF11}, it is expected, but not certain, that the effect of too much source task training would decrease with increasing target task training size.

\subsubsection{Does source task training set size affect loss of transfer accuracy associated with too much source task training?}

We also experimented with the CalTech101 dataset. It is both much smaller than either of the other datasets and more unbalanced. It has an approximate median of 50 SPC, as opposed to 500 SPC, which CIFAR100 and Tiny ImageNet200 have. Furthermore, Caltech101 has classes with as few as 28 samples and  classes with as many as a few hundred samples.  This led to different behavior than seen with the other two datasets.
 
To determine if this difference was due to the decreased size  (i.e. decreased amount of training per epoch) of the Caltech101 dataset, we took the source task training set from the Tiny ImageNet200 dataset and abridged it to 50 SPC. Then, in order to increase the size of this training set (i.e. recreate an equivalent amount of training per epoch), without changing the variety of data, we copied  each image 10x, giving us 500 SPC again.

\subsubsection{Does learning rate affect loss of transfer accuracy associated with too much source task training?}

There was a weak correlation between optimal target task performance and decreases in learning rate. This raised the question of whether the decline in target task learning was caused by the amount of training that had occurred at the time of the learning rate decreases, or by the decrease itself. So, we performed an experiment with the WRN-28-10 architecture and CIFAR100 dataset, using delayed learning rate decreases at 100, 160 \& 180 epochs instead of at 60, 120 \& 160 epochs.

\section{Results}

Each of our graphs shows median source task accuracy with a green curve that uses the right y-axis. The target task results use the left y-axis. Transfer learning  results are indicated with a data point for the median accuracy and `error bars' to indicate the max and min values, obtained over 25 trials. Red denotes target task results obtained with 0 source task training epochs - i.e. no transfer learning. Yellow indicates target task results using the net with optimal source task performance (i.e. the nets currently used for transfer). Because each of the 5 source nets achieved optimal performance at a different epoch, we associate the target task results for the epoch of peak source task performance with the median of those epochs. Dark blue indicates the best target task results obtained without using the  optimal source task performance net. Horizontal  lines are used to aid comparison of the median accuracies. 

\subsection{Specific Experimental Results}
\subsubsection{ Does optimal source task performance imply optimal pre-training for a target task?}
The results obtained in Figs. \ref{fig:WRN_VGG} \& \ref{fig:Tiny_Caltech}, clearly show that optimal source task performance does not imply optimal pre-training for a target task.  Fig. \ref{fig:WRN_VGG} shows this for the CIFAR100 dataset with both a WRN-28-10 and VGG architectures, and for 10, 50 and 250 SPC in the target class training set. Fig. \ref{fig:Tiny_Caltech} shows the same result for the TinyImagenet200 and Caltech101 datasets, using a WRN-28-10 architecture and 10 SPC in the target class training set. The difference in accuracy reflected by the dark blue horizontal line (i.e. best overall transfer accuracy) and the yellow horizontal line (i.e. accuracy achieved by pre-training to optimal source task performance) indicates how much training to peak source task performance may degrade transfer learning of the target task from training to the optimal stopping point.

\subsubsection {Does too much source task training place the pre-trained net in a state that is overly specific to relearning the source task?}
\begin{figure}
    \centering
    \hfill
    \begin{subfigure}{\linewidth}
       \includegraphics[width=.5\columnwidth]{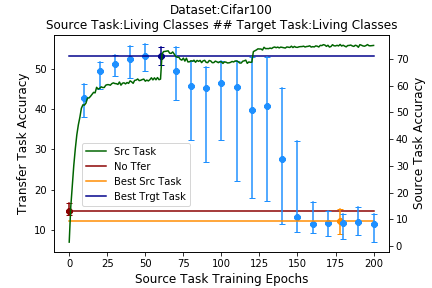}\hfill
       \includegraphics[width=.5\columnwidth]{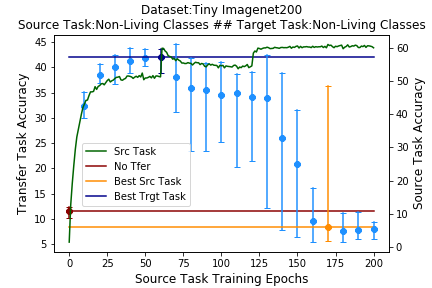}
    \end{subfigure}
    \hfill
    \caption{Transfer learning of the original source task, using a WRN-28-10 architecture and 10 SPC. The left graph shows this using the living classes of the CIFAR100 dataset as source and target task. The right graph does so for the set of non-living classes from the TinyImagenet200 dataset.}
    \label{fig:source_task_tfer}
\end{figure}

To test whether our nets had been pre-trained to a state that was overly specific to their source task, we attempted transfer learning with \textbf{identical} source and target tasks, using 10 SPC for target task training, a WRN-28-10 architecture and both the CIFAR100 and TinyImagenet200 datasets. The results shown in Fig. \ref{fig:source_task_tfer} demonstrate that neither net is in a state that is overly specific to the source task.  In each case, not only is the original optimal source task net far from optimal at relearning that \textbf{same} task, but it also, surprisingly, displays negative transfer.

%repeated the experiments from Figs. \ref{cifar100_l_nl} and \ref{tiny200_nl_l}, but this time we attempted to relearn the original source task with transfer learning, in a `degenerate' example of transfer learning. The results in Figs. \ref{cifar100_l_l} and  \ref{tiny200_nl_nl} demonstrate that   the net is not in a state that is overly specific to the source task.  In each case, not only is the original optimal source task net far from optimal at relearning that same task, but it also, surprisingly, displays negative transfer.

Negative transfer is evidenced by the fact that the horizontal yellow line (i.e. target task accuracy achieved after obtaining peak source task accuracy) is beneath the horizontal red line (i.e. target task accuracy achieved without any source task training). 

The loss of the pre-trained net's accuracy, when relearning the original source task, demonstrates that excessive training on a source task can leave a pre-trained net in a state where it is less able to learn. This is similar to the `fragile co-adapted' nodes first observed by Yosinsky et al.\cite{YosinskiCBL14}, except we've demonstrated that  `fragile co-adaptation' can occur at the high, rather than middle, layers of a net and that this phenomenon can be associated with the amount of source task training.

\subsubsection{Does target task training set size affect loss of transfer accuracy associated with too much source task training?}
The results in Fig. \ref{fig:WRN_VGG} show that as the target task training size increases, not only do the accuracy benefits of transfer learning decrease, as expected, but also the relative importance training for the optimal number of source task training epochs also decreases.

\subsubsection{Does source task training set size affect loss of transfer accuracy associated with too much source task training?}
\begin{figure}
    \centering
    \begin{subfigure}{\linewidth}
       \includegraphics[width=.5\columnwidth]{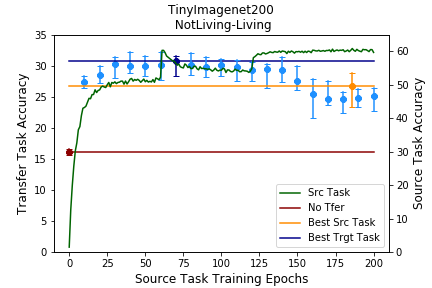}\hfill
       \includegraphics[width=.5\columnwidth]{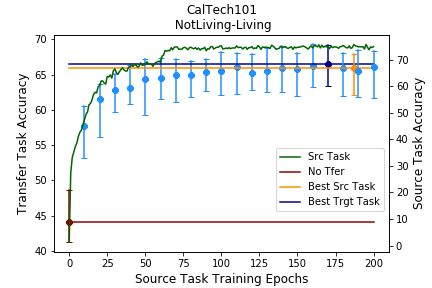}
    \end{subfigure}
    \caption{Transfer learning from set of living things to set of non-living things, using a WRN-28-10 architecture, 10 SPC, and the TinyImagenet200 and Caltech101 datasets, respectively.}
    \label{fig:Tiny_Caltech}

\end{figure}

The results obtained using the Caltech101 dataset (right side graph of Fig. \ref{fig:Tiny_Caltech}) with 10 SPC for the target task training set and a WRN-28-10 architecture were far less significant than those obtained under the same conditions with TinyImagenet200 (left side graph of Fig. \ref{fig:Tiny_Caltech}) or with the CIFAR100 dataset for both the WRN-28-10 and VGG architectures (left most graphs in Fig. \ref{fig:WRN_VGG}). We observed that the Caltech101 dataset is notably smaller and less balanced than CIFAR100 and Tiny ImageNet200. The smaller size means that not only does Caltech101 have less variety than our other two datasets, but it also engages in less updating per training epoch. 

\begin{figure}
    \centering
    \begin{subfigure}{\linewidth}
       \includegraphics[width=.5\columnwidth]{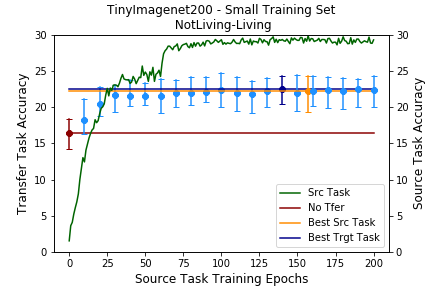}\hfill
       \includegraphics[width=.5\columnwidth]{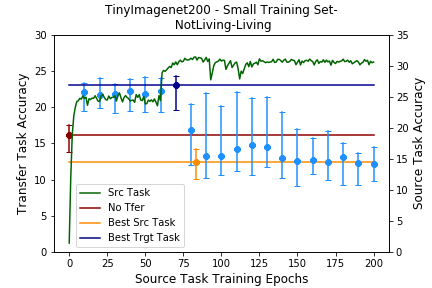}\hfill
    \end{subfigure}
   \caption{Graphs examining the effects of small source task training sets on transfer. Left graph artificially shrinks source task training set to 50 SPC. Right graph does same, but repeats each sample 10X. Both of these experiments used WRN-28-10 architecture.}
   \label{fig:small}
\end{figure}

To investigate the degree to which the smaller size of the Caltech101 dataset was the cause of this phenomena, we truncated the source task data from Tiny Imagenet200 dataset to 50 SPC. We felt this would help it to best mimic the Caltech101 dataset. The transfer learning experments were then repeated, with the results shown in the first graph of Fig.\ref{fig:small}. It shows the same lack of degradation of transfer learning as the right side graph of Fig. \ref{fig:Tiny_Caltech} .

The middle graph of Fig. \ref{fig:small} shows results when each image in the source training set of our truncated Tiny ImageNet200 is repeated 10x each epoch. This restores the size of the original training set, though not its variety. As can be seen, the degradation in the transfer task accuracy reappears, in a fairly extreme fashion, showing that it is more due to excessive training and not very influenced by variety in the dataset. Even more significantly, it shows evidence of negative transfer, for the net with optimal source task performance, since the horizontal yellow line has fallen below the horizontal red line.

\subsubsection{Does learning rate affect loss of transfer accuracy associated with too much source task training?}

\begin{figure}
    \centering
    \begin{subfigure}{\linewidth}
       \includegraphics[width=.5\columnwidth]{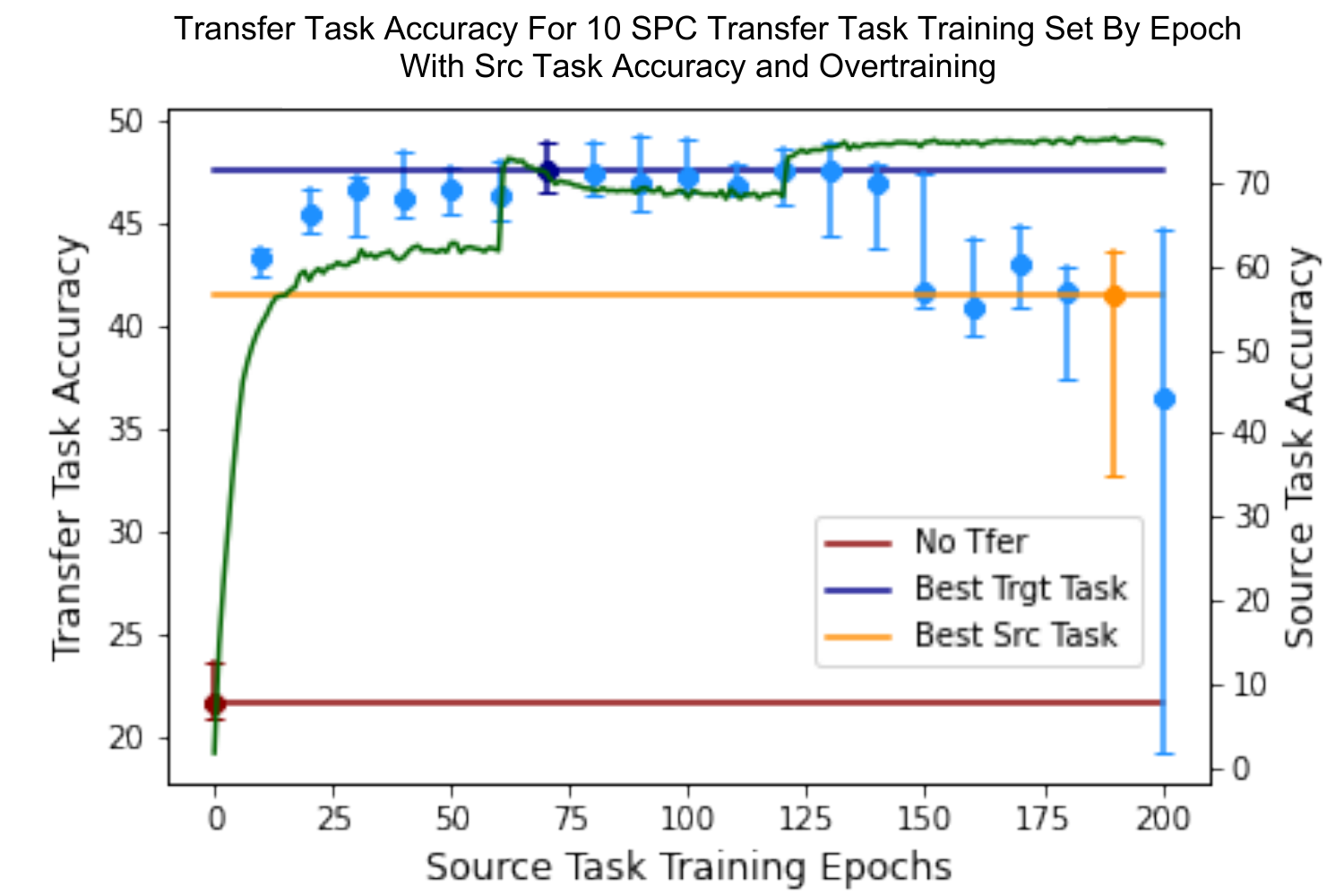}\hfill
       \includegraphics[width=.5\columnwidth]{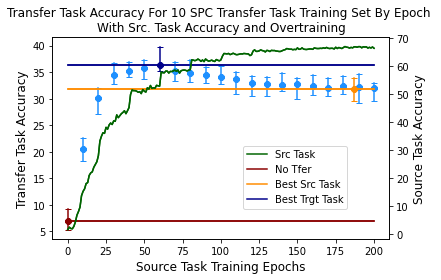}\hfill
       \caption{10 SPC: WRN-28-10 architecture(left); VGG16 architecture(right)}
    \end{subfigure}
    \begin{subfigure}{\linewidth}
       \includegraphics[width=.5\columnwidth]{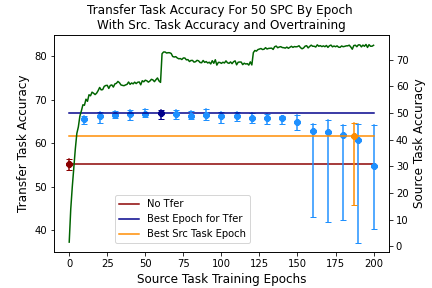}\hfill
       \includegraphics[width=.5\columnwidth]{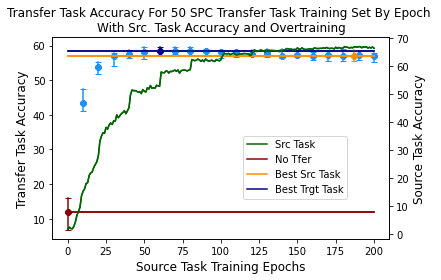}\hfill
       \caption{50 SPC: WRN-28-10 architecture(left); VGG16 architecture(right)}
    \end{subfigure}
    \begin{subfigure}{\linewidth}
       \includegraphics[width=.5\columnwidth]{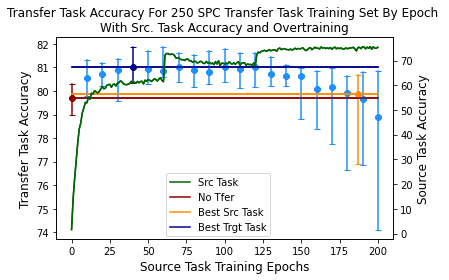}\hfill
       \includegraphics[width=.5\columnwidth]{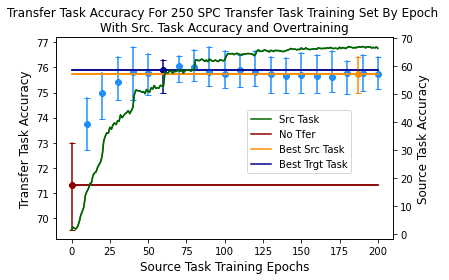}\hfill
       \caption{250 SPC: WRN-28-10 architecture(left); VGG16 architecture(right)}
    \end{subfigure}
    \hfill
    \caption{Transfer learning with WRN-28-10 Architecture and VGG16 Architecture from set of living things to set of non-living things using CIFAR100 data and 10, 50 and 250 SPC, respectively.}
    \label{fig:WRN_VGG}
\end{figure}

\begin{figure}
    \centering
    \begin{subfigure}{\linewidth}
       \includegraphics[width=.5\columnwidth]{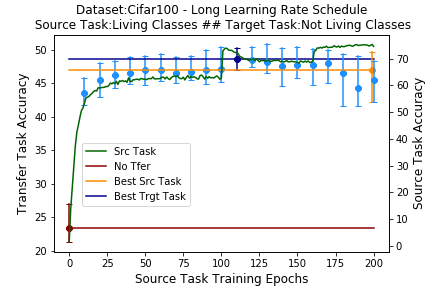}
    \end{subfigure}
    \caption{Graph examining effects of learning rates on over source task training by repeating earlier experiments with a longer learning rate schedule. This experiment used WRN-28-10 architecture.}
    \label{fig:learn_rates}
\end{figure}

Fig. \ref{fig:learn_rates}  illustrates the importance of learning rate, as opposed to just number of training epochs. It repeats the first experiment from  Fig. \ref{fig:WRN_VGG}  using delayed learning rate decreases at 100, 160 \& 180 epochs instead of at 60, 120 \& 160 epochs. The shift in occurrence of the decreases in transfer learning accuracy to coincide with the scheduled decrements in learning rate, indicates the effect smaller learning rates have in creating pre-trained nets that are ill-suited for transfer.

\section{Conclusion}
Fine tuning of pre-trained nets is one of the most common transfer learning techniques for 
deep nets. We have demonstrated that optimal source task performance is not indicative of optimal target task performance for pre-trained nets. Too much source task training results in a pre-trained net that is less able to learn target tasks. This is consistent with the observation that if training continues beyond a certain point, a net becomes less sensitive to the error signal produced by the training data, which renders it less capable of continued learning\cite{achille2017critical, shwartz2017opening, chaudhry2018riemannian,  saxe2019information, ash2020warm, berariu2021study}. It also bears great similarity to the `fragile co-adapted features' first noted by Yosinski et al.\cite{YosinskiCBL14}, except we observe them in the highest layer, which is commonly used for pre-training. As a result,  more caution is warranted in the use of pre-trained nets, especially for small target task training sets.

\clearpage
\section*{Acknowledgment}
The authors would like to thank Travis Cuvelier for many useful conversations.

\ifCLASSOPTIONcaptionsoff
  \newpage
\fi



%\bibliographystyle{IEEEtran}
\bibliography{./refs}
\bibliographystyle{IEEEtran}

\end{document}